# The yield curve as a recession leading indicator. An application for Gradient boosting and Random Forest.

Pedro Cadahia Delgado, Emilio Congregado,[1] Antonio A. Golpe[1], José Carlos Vides[2]

Department of Economics. University of Huelva, Plaza de la Merced, 11, 21002, Huelva, Spain.

Department of Applied and Structural Economics and History, Faculty of Economics and Business, Complutense University of Madrid, Madrid, Spain.

## ABSTRACT

Most representative decision-tree ensemble methods has been used to examine the variable importance of Treasury term spreads to predict US economic recessions with a balance of generating rules for US economic recession detection. A strategy is proposed for training the classifiers with Treasury term spreads data and compare the results in order to select the best model for interpretability. We also discuss the use of SHapley Additive exPlanations (SHAP) framework to understand US recession forecasts by analyzing feature importance. Consistently with the existing literature we find the most relevant Treasury term spreads for predicting US economic recession and a methodology for detecting relevant rules for economic recession detection. In this case, the most relevant term spread found is 3-month–6-month, which is proposed to be monitoring by economic authorities. Finally, the methodology detected rules with high lift on predicting economic recession that can be used by these entities for this propose. This latter result stands in contrast to a growing body of literature demonstrating that machine learning methods are useful for interpretation comparing many alternative algorithms and we discuss the interpretation for our result and propose further research lines aligned with this work.

## I. INTRODUCTION

Since the decade of the '80s, economic crises have been more recurrent and deeper. In this respect, researchers and practitioners have tried to understand, model, and even predict a recession differently. One popular forecasting tool suggested in the literature and followed by economists is the analysis of the slope of the yield curve or the term spread, i.e., the difference between long-term and short-term interest rates [1].

According to this idea, in a competitive financial environment, the term structure should respond to international market forces, considered as key for assessing the impact of monetary policy and more importantly, to express the economy's behaviour. Indeed, if a monetary policy is effective, changes in short-term policy interest rates should impact long-term ones [2]. In this sense, the need to forecast and prevent economic recessions has become of great importance to policymakers, practitioners and researchers. In this respect, the use of economic and financial variables as predictive information containers joint to the application of several econometric methods and machine learning models have focused on detecting a better accuracy in predicting the possible turning points of the business cycle and, more deeply, economic recessions [3]. This literature review has tried to shed some light on the more important and highlighted topic works.

As previously mentioned, the term structure holds implications in macroeconomics or finance and the shape of the yield curve (see [4] for a survey). According to this, an upward sloping yield curve suggests that future short-term rates are expected to rise. Contrariwise, a descending sloping yield curve may mean that future short-term rates are expected to drop. Like [5] state, the yield curve's slope –the difference between the longer maturity of interest rates and the shorter maturity– gives an important source of information of the real economy evolution. Accordingly, they found that a positive curve slope is associated with future increases in real economic activity when using macroeconomic variables, possessing a significant predictive power or its economic implications in the monetary policy [6] and [7]. To understand the background of the term structure, we briefly treat the Expectations Hypothesis of Term Structure (EHTS). This hypothesis illustrates the relationship between short and long-term interest rates and represents the most influential theory explaining the term structure relations. This hypothesis establishes that long-term interest rates are defined by an average of the contemporary and expected short-term interest rate [8]. Therefore, this relationship between both types of interest rates indicates that their spread holds meaningful information on future changes in short-term rates and is an important function in the potential effectiveness of monetary policy [9] and [10] or reflecting economic agents' anticipations of future events such as recessions, for instance (see [11] for a survey). According to [12], the inversion of the yield curve is viewed as a consistent predictor of recessions and future economic activity, providing an important reason to explain the flattening or inversion of the yield curve: a monetary lightening. A tightening monetary policy would be considered a rise in short-term interest rates, focusing on reducing inflation. The consequence of the monetary tightening is that the economy may slow down.

Consequently, shorter-term interest rates are considered indicators of demand for credit and future inflation. Therefore,

longer-term interest rates would tend to decrease and flatten the yield curve, an example of the relation between the yield curve behaviour and recessions. Definitely, the yield curve's steepness would help us predict and determine a future recession [13].

The literature on this topic has tried to demonstrate the role of the term structure or the yield curve as a good forecasting tool for recessions [14]. The influential papers of [5] and [15] should be noted. These works evidenced that the yield curve might be employed to predict real growth in consumption, investment, or aggregate GNP, and more importantly, they demonstrated the relation with NBER-dated recessions. For its part, [16] suggests that among different variables used in his work, the term spread is the significant predictor of recessions at horizons beyond three months. In this respect, many previous papers have treated the topic by relating the GDP growth with the yield curve slope (see [17]-[25], among others or [26] for a deep survey of the topic.). Another important work by [27] argues the convenience of applying models which use the yield curve to predict recessions. In other influential papers in the literature, the term spread is also useful in predicting recession even for professional forecasters, as [28] suggested and [29] combined the term spread with stock returns to measure the accuracy of the term spread the latter to predict recessions. His results were positive, and the term spread was found as a valuable predictor of recessions for German and US economies. In a similar work by [28], [30] compared the strength of the yield curve in forecasting recessions with the data used in [28], evidencing the power of the former and suggesting the suitability of using this indicator. For its part, [31] also treated the capability of predicting recessions of the term structure and highlighted the power of this indicator over other leading indicators. Its strength decreased as a predictor after the financial crisis due to the volatility of macroeconomic variables, but unfortunately, its predictive power over the last decade has fallen.

Furthermore, [3] in line with the previous literature, find that the ability of the term structure to predict recessions is stronger over the twelve-month horizon when using a similar probit model than [5] or [13] used. Additionally, [32] further evidenced the potential of the yield curve in forecasting future situations of the US economy over horizons ranging from one quarter to two years. Besides, [33] recognized that the yield curve contains information on future GDP growth and that its predictability varies with time, forecast horizons, and quantiles of the distribution of future growth; nonetheless, a significant empirical contribution of their work is that it seems more efficient to predict future expansionary phases, which are more common than recessions, for which the latter appears to perform better. Finally, although [34] find that developments in the stock market diminish the efficacy of the yield curve in forecasting future economic activity, they show the fitness of this indicator for predicting economic activity in many most important world economies, such as the US, Canada and Europe and, more importantly, when periods of financial stress are analyzed.

From another empirical perspective, it emerges in the literature the use of techniques based on machine learning algorithms. In this sense, [35] claims the suitability of machine learning techniques on central banking or monetary policy issues as applied in other real-life topics. In this sense, [36] demonstrated the yield curve as a robust and consistent predictor of economic activity when US business cycle turning points are checked by using four different methods, i.e., equally-weighted forecasts, Bayesian Model Averaging (BMA), and linear and non-linear machine learning boosting algorithms. An important paper in the literature by [37] compares different Support Vector Machine (SVM hereafter) and logit models when using the yield curve as a leading indicator, being "the first empirical investigation on the relation between the yield curve and an economy's real output, using an SVM classifier". The model created is helpful for policymakers in order to forecast future recessions. In order to reaffirm this latter study, [38] the yield curve is a useful tool for assessing future economic activity, achieving a 100% forecasting accuracy for recessions. For its part, [39] demonstrated that the predictive power of boosted regression trees is considerably better than standard probit models. Their findings show that short rates and the yield curve are crucial leading indicators for recession forecasts during the 1974-2014 period. Finally, [40] employ several machine learning methods such as Least Absolute Shrinkage and Selection Operator (LASSO), and Elastic Net, Discriminant Analysis classifiers, Bayesian classifiers, and classification and regression trees (CART), in line with the existing literature and reveal the ability of the yield curve to act as an early warning system to predict recessions in the United States is reconfirmed. Specifically, the yield curve keeps on a consistent and reliable predictor of recession over the 12-month forecast horizon and [41] also apply a battery of machine learning methods: decision trees, random forests, extremely randomized trees, support vector machines (SVM), and artificial neural networks, finding that almost all the machine learning models appropriately predict the global financial crisis of 2007-2008 and, additionally, they indicate that the flatter or more inverted the yield curve is, the higher the chance of a crisis, exposing the tendency of chasing performance or increased risk-taking that can often be seen before financial crises.

To the best of our knowledge, our approach, i.e., Gradient Boosting and Random Forest Machine Learning methods, allows us to reach a better accuracy than in those previous papers on the topic. These Machine Learning algorithms let us identify the more relevant variables associated with the main variable, which has not been done before in the literature. Additionally, we extend the time horizon, i.e., we update data compared to previous studies. Indeed, our results indicate that our algorithm let us signal and choose the most influential variables for predicting economic recessions amongst the term spreads analysed. This case highlights some of the most important term spreads as 3-month–6-month, 2-year–5-year and 5-year–10-year. Furthermore, concerning these variables, the lift metric is computed to detect intervals with a higher probability of accounting for a recession, applied to the rules description methods. Results suggest that the most important term spread is 3-month–6-month compared with the term spreads mentioned in the literature. Results give some considerations for monetary authorities, policymakers and practitioners, such as the monitorisation of this term spread above mentioned as a tool for evidencing economic recessions.

The rest of the paper is as follows. Section II presents the data and methodology used in the paper. Later, section III show and discuss the results; the concluding remarks are in section

*IV.DATA AND METHODOLOGY*

## A. Introduction

A supervised method is proposed to predict economic crisis cycles and can also identify the key factors that lever this phenomenon. Assessing variable importance is an important task; this is reflected in many studies fields; besides, several approaches address this question [42]-[45].

A decision-tree ensemble classification method is proposed for interpretability rather than only predicting economic recessions from the different term spread as independent variables. In this way, the variable importance is computed to measure which variables are the most relevant to predict economic crisis cycles. More interpretation of the model is performed by analyzing the dependencies with the most correlated variables and the feature value dependency regarding the target variable to understand this phenomenon better. Finally, a rule extraction process is proposed that could be useful for interpreting and detecting economic recession.

## B. Data Description

For our empirical analysis, we employ a monthly sample of Treasury Constant interest rates at nine different maturities from January 1969 to November 2020 (amounting to 601 observations for each interest rate series). The data corresponds to the constant maturity rates of 3-month, 6-month, 1-year, 2-year, 3-year, 5-year, 7-year, 10-year and 20-year.

The data is collected from the Federal Reserve Economic Data (FRED) collected by the Economic Research Division of the Federal Reserve Bank of St. Louis. Since the 1-month Treasury Constant maturity rate is only accessible since January 2001, we have picked these maturities considering the availability of consistent interest rate data with the period studied. We reveal 3-month, 6-month and 1-year as short-run, including the latter variable 1-year as short-term because it offers more robustness in our assessment. Conversely, we contemplate the rest of the maturity rates as long term. Table I shows descriptive statistics related to each interest rate in different maturities. These variables show similar behaviour in terms of volatility, and Fig.1 A and B presents a plot analysis of the time series traced for all maturities.

TABLE I. Descriptive statistics for the data

|        | M3   | M6   | Y1   | Y2   | Y3   | Y5   | Y7   | Y10  | Y20  |
|--------|------|------|------|------|------|------|------|------|------|
| Mean   | 4.57 | 4.69 | 5.08 | 5.18 | 5.54 | 5.84 | 6.07 | 6.23 | 6.31 |
| Median | 4.86 | 4.95 | 5.27 | 5.03 | 5.77 | 5.97 | 6.17 | 6.20 | 6.01 |
| Min    | 0.01 | 0.04 | 0.10 | 0.13 | 0.16 | 0.27 | 0.56 | 0.62 | 1.06 |
| Max    | 16.30 | 15.52 | 16.72 | 16.46 | 16.22 | 15.93 | 15.65 | 15.32 | 15.13 |
| SD.    | 3.41 | 3.40 | 3.64 | 3.78 | 3.51 | 3.53 | 3.23 | 3.11 | 3.05 |

[a] Data from January of 1969 to November of 2020.
[b] M and Y refers to month and year respectively.

From 9 interest rates, 36 spread variables are obtained, the calculation being a subtraction of two elements; this follows a combination without repetition C(n,r), being n and r the set and subset size, respectively. As shown in Table I, the interest rates show similar statistical properties. Nevertheless, the short term interest rates 3-month and 6-month presents lower mean and median and higher standard deviation. On the contrary, long term interest rates show the opposite higher mean and median and lower standard deviation. Henceforth for representing term spread at figures and tables, due to saving space, an abbreviation is used, being M and Y for month and year interest rates respectively, i.e. M3-Y10 for 3-month–10-year term spread.

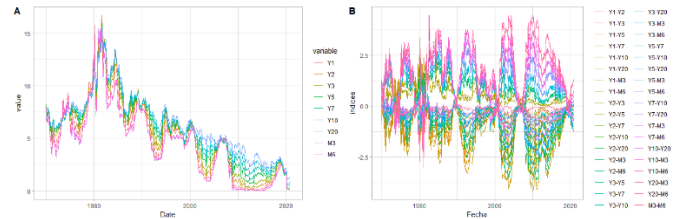

Fig. 1. Original data interest rates(A) & Computed Term spreads(B).

At Fig.1 A, the interest rates are plotted where the general trend is decreasing, Fig.1 B shows the computed Term spread for all combinations of interest rates, it is stated that there are some expansion stages with the behaviour of divergence and flattening stage where the term spreads are inverted with the behaviour of convergence which could be an early indicator of economic recession.

As a combinatory result, the term spread variables show several strong correlations. The correlation coefficient is used to verify collinearity, and it is argued that collinearity is certain at the 0.9 level of a correlation coefficient or higher [46]. A correlation analysis is shown between variables at Fig 2, where the correlation plot shows the coefficients:

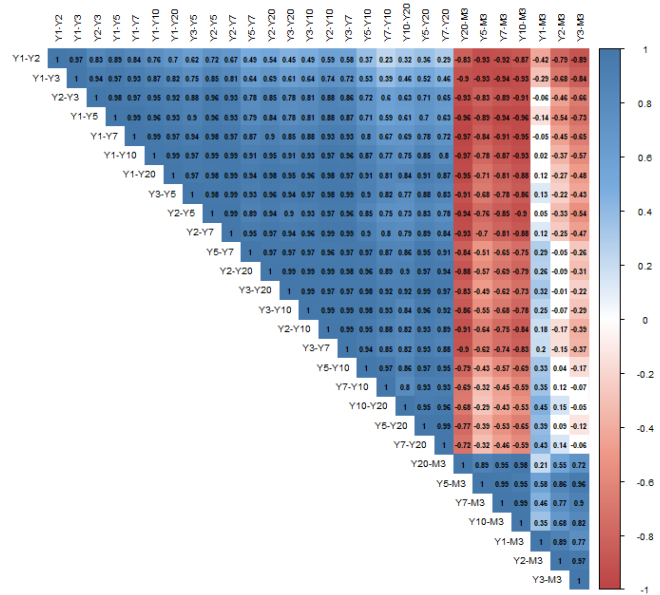

Fig. 2. Pearson correlation between term spread variables.

Pearson's correlation results in Fig.2 shows high correlated features. In line with the literature, results show a consistent negative relationship in the difference between long-term and short-term interest rates and consequently in the term spreads [1]. This is taken into account to interpret the importance of the features exposed in the results.

Literature mainly focused on continuous variables whose values, for instance, growth rates in GNP, GDP, industrial production, consumption, investment, among others [1]. In this work, only interest rates are used as predictors as the main purpose of this work is not to offer the better predictive model results of literature but to understand the relationships, importance and rules

* Corresponding authors:
E-mail address: jvides@ucm.es (Jose Carlos Vides)

regarding interest rates with an economic recession.

*Variable Target Lift*

Regarding materials, you should include a description of examined objects and tools used during the experiment. Give every detail that could affect experiment results.

In machine learning, Lift is a metric used to assess the performance of a targeting model at predicting or classifying cases as having an enhanced response concerning the population as a whole.

This metric is pretty straightforward to understand, and a targeting model is performing well if the response within the target is much better than the average for the population. In other words, Lift is simply the ratio of these values: target response divided by average response [47]. It is defined as:

$$Lift = \frac{P(A \cap B)}{P(A)P(B)} \quad (1)$$

These indicators, shown in Table II, are useful in the exploratory data analysis stage to understand at each variable's decile which range of values of the response variable has more impact on positive target. This can be used as an early exploratory rule for detecting economic recession, and this is complementary information as the decile split does not guarantee the optimal value range for a variable for maximizing the lift; on the contrary, the computed lift for tree base rules ranges may give a better separation as it is a supervised method, for this reason, it helps initially to understand this economic processes.

TABLE II. Lift for crisis per Deciles for the most relevant features.

| Decile | M3-M6 | Y3-M3 | Y5-Y10 | Y2-Y5 | Y2-M6 | Y3-Y7 |
|---|---|---|---|---|---|---|
| 1 | **1.46** | **1.09** | 0.46 | 0.16 | **1.52** | 0.33 |
| 2 | 0.74 | **1.60** | **1.14** | 0.91 | 0.62 | 0.87 |
| 3 | **1.20** | 0.75 | 0.90 | **1.40** | 0.00 | **1.11** |
| 4 | 0.51 | 0.53 | 0.64 | **1.67** | 0.91 | 0.96 |
| 5 | 0.85 | **1.42** | **1.63** | **1.55** | **1.71** | **1.26** |
| 6 | 0.77 | **1.29** | 0.56 | 0.78 | **1.71** | 0.62 |
| 7 | 0.34 | **1.42** | 0.31 | 0.62 | 0.62 | **1.09** |
| 8 | 0.41 | 0.66 | 0.71 | **1.26** | 0.30 | 0.33 |
| 9 | **1.88** | 0.54 | 0.62 | 0.30 | 0.78 | **1.42** |
| 10 | **1.79** | 0.75 | **2.95** | **1.34** | **1.83** | **2.04** |

a Term spread abbreviations contains M and Y for monthly term and yearly term interest rates respectively.

For the sake of simplicity, in Table II the target lift is computed only for the most important variables, as shown in section III. From this table, some initial patterns there can be found. Generally, almost every term spread at high deciles has a high lift in economic recession except for 3-year–3-month. On the contrary, the 3-year-3–month and 2-year–6-month term spreads show high lift for low and mid deciles. This is an initial indicator due to the higher probability of recession in those deciles; for specific range values, the decile's interval table can be found in the appendix.

C. Methodology

The main purpose of this work is not only to offer a model for predicting economic recessions but also to offer a methodology of a good enough model that is able to explain variable importance, dependencies and economic recession detection rules.

Decision-tree ensemble methods are supervised learning methods for modeling the relationship between the dependent variable y with the characteristic vector x. Besides, these techniques are a common choice on the actual machine learning research scenario, it has a wide range of applications for regression, classification and other tasks [48], [49].

The two main decision-tree ensemble methods in bagging and boosting for classification scenario are applied in this work for estimating the economic crisis cycles. The advantage of this methods are that often provides predictive accuracy that cannot be beat, it can optimize on different loss functions and provides several hyperparameter tuning options that make the function fit flexible, generally no data pre-processing required and often works great with categorical and numerical values.

To train the models, a training and test data split is performed, where the training set consist on all available variables for all observations from January of 1969 to December of 1999 and the test set comprises from January of 2000 to January of 2020, with the correspondent binary supervised target of economic crisis cycle. In other words, the models should learn which features are relevant in order to predict from an time interval selected for another more recent time interval which should be relevant not only for predicting the economic crisis cycles but also for Interpretability of the actual situation.

*Random Forest Classifier*

Random Forest (RF) was proposed by [50] as an ensemble method for regression based on individual decision trees, the original classification approach based on Stochastic Discrimination was proposed by [51], [52].

In this way, Ranger is a fast implementation of RF [53] or recursive partitioning, particularly suited for high dimensional data. The R implementation Ranger was used to adjust a RF model respectively the considered optimal settings [54].

Which makes Random forest powerful is that builds several weak decision trees in parallel, resulting computationally cheap process, by combining the trees to form a single, strong learner by averaging or taking the majority vote results often to be accurate learning algorithms.

The pseudocode is illustrated at Algorithm scheme 1. The algorithm works as follows: for each tree in the forest, a bootstrap sample is selected from S where S(i) is the ith bootstrap. Then it is trained a decision-tree as follows: at each node of the tree, instead of examining all possible feature-splits, a random features subsect selection is made f ‚äÜ F. where F is the set of features. The node then splits on the best feature in f rather than F. In practice f is much, much smaller than F. By narrowing the set of features, it drastically speed up the learning of a tree.

| ALGORITHM I. Random Forest algorthm |
|---|
| **Precondition**: A training set $S \coloneqq (x_1, y_1), \ldots, (x_n, y_n)$, being $F$ the features and $B$ number of trees in forest. |
| 1      **function:** RandomForest($S, F$) |
| 2          $H \leftarrow \emptyset$ |
| 3          **for** $i \in 1, \ldots, B$ **do**: |
| 4              $S^{(i)} \leftarrow A\ boostrap\ sample\ from\ S$ |
| 5              $h_i \leftarrow RandomizedTreeLearn(S^{(i)}, F)$ |
| 6              $H \leftarrow H \cup \{h_i\}$ |
| 7          **end for** |
| 8          return H |
| 9      **end function** |
| 10     **function** RandomizedTreeLearn($S, F$) |
| 11          **At** each node: |

| 12 | $f \leftarrow$ small subset of $F$ |
| 13 | Split on best feature in $f$ |
| 14 | **return** learned tree |
| 15 | **end function** |

RF algorithm is a bagging technique for building an ensemble of decision trees, and this technique is known to reduce the variance of the algorithm. Traditionally bagging with decision trees, the constituent decision trees may be highly correlated because the same features will tend to be used repeatedly to split the bootstrap samples. At the same time, restricting each split-test to a small, random sample of features decreases the correlation between trees in the ensemble and improves the performance of the algorithm.

### 1. Gradient Boosting Machine

The gradient boosting machines (GBM) proposed by [55] is a robust machine learning algorithm due to its flexibility and efficiency in performing regression tasks [55].

The main difference between boosting and traditional machine learning techniques is that optimization is held out in the function space. In other words, the function estimate f̂ is parametrized in the additive functional form:

$$\hat{f}(x) = \hat{f}^M(x) = \sum_{i=0}^{M} \hat{f}_i(x) \qquad (2)$$

In this notation, M is the number of iterations, $\hat{f}_0$ is the initial guess and $\{\hat{f}_i\}_{i=1}^{M}$ are the function increments, also known as "boosts".

To ensure that the functional approach is achievable in practical terms, a comparable approach to parameterization of the family of functions can be implemented. It is introduced to the reader the parameterized "base-learner" functions $h(x, \theta)$ to differentiate it the overall ensemble functions estimates $\hat{f}(x)$. Different families of basic learners can be chosen, such as decision trees and loss functions.

The "greedy stagewise" approach of function incrementing with the base-learners can be formulated.

For the function estimate at the t-th iteration, the optimization function is:

$$\hat{f}_t \leftarrow \hat{f}_{t-1} + \rho_t h(x, \theta_t) \qquad (3)$$

$$(\rho_t, \theta_t) = \arg\min_{\rho, \theta} \sum_{i=1}^{N} \psi(y_i, \hat{f}_{t-1}) + \rho h(x, \theta_t) \qquad (4)$$

The optimal step-size $\rho$, should specified at each iteration.

The gradient boosting algorithm proposed by Friedman[55], can be summed up with the following pseudocode at algorithm 3.

| ALGORITHM II. Friedman's GBM algorithm |
|---|
| **Precondition:** |
| • Input data $(x, y)_{i=1}^{N}$ |
| • Number of iterations M |
| • Choice of loss-function $\Psi(y, f)$ |
| • Choice of the base-learner model $h(x, \theta)$ |
| 1    Initialize $\hat{f}_0$ with a constant |
| 2    **for** t=1 to M **do**: |
| 3        compute the negative gradient $g_t(x)$ |
| 4        fit a new base-learner function $h(x, \theta_t)$ |
| 5        find the best gradient descent step size $\rho_t$: |
|        $\rho_t = \arg\min_{\rho, \theta} \sum_{i=1}^{N} \psi(y_i, \hat{f}_{t-1}) + \rho h(x, \theta_t)$ |
| 6        update the function estimate: $\hat{f}_t \leftarrow \hat{f}_{t-1} + \rho_t h(x, \theta_t)$ |
| 7    **end for** |

The theory and formulation of GBM are available in reference [55], which interested readers in a more profound explanation for a better understanding of this method.

In this work, the so-called Extreme Gradient Boosting Training(XGB), proposed by [56], a version of GBM, was applied as a boosting method for classification with the R library xgboost.

### 2. Classifier Evaluation

For training the model, a data partition was performed; as explained in the previous sections, the predictive accuracy of the models was measured by splitting the data into training and test sets.

The training set comprehends from 1970 to 1999 with 360 instances and a binary target variable with 16% positives(5 crisis cycles). The test set comprehends from 2000 to 2020, which are 251 instances with 14% of positives in the binary target(3 crisis cycles).

As a classification task, the error assessment was performed using the predicted class for the selected models and computing some accuracy metrics from the confusion matrix.

Let {P, N} the positive a negative instance class and let $\{\tilde{P}, \tilde{N}\}$ be the predictions produced by a classifier. Let P(P|I) be the posterior probability that an instance I is positive.

TABLE III. Classification Metrics for classification model assessment.

| Metric | Formula |
|---|---|
| Recall(TPR) | $P(\tilde{P}\|P) \approx \dfrac{positives\ correctly\ classified}{total\ positives}$ |
| Specificity(TNR) | $P(\tilde{N}\|N) \approx \dfrac{negatives\ correctly\ classified}{total\ negatives}$ |
| Precision(PPV) | $\dfrac{positives\ correctly\ classified}{positives\ correctly\ classified + Negatives\ correctly\ c}$ |

There is no unique metric for assessing a classification task, depending on the characteristics to be evaluated, we consider precision as the most suitable metric for this purpose as considers the positives correctly classified within the observations correctly classified.

### 3. Model Interpretation

The interpretability of a statistic model helps to understand why certain decisions or predictions have been made; for this reason, measuring variable importance is an important task in many applications. In this sense, this is the era of making machine learning explainable; several authors have conducted an extensive review of methods [57, 58].

The most common variable importance based has been tested by several researchers using both simulated and real data; this metric tends to be biased in many scenarios [58]-[60]. As studied in subsection II.B., there is the presence of mutually correlated and collinearity; Gini variable importance is expected to be biased [59],

[60].

Nevertheless, there is also another classification for interpretability, and it could be either local or global; in other words, it is explaining an individual prediction or the entire model behaviour [61].

### a. SHAP Variable Importance

SHapley Additive exPlanations(SHAP) is a model additive explanation approach in which each prediction is explained by the contribution of the features of the dataset to the model's output[62], [63]. SHAP comes from the game theory field, that is, the solution for the problem of computing the contribution to a model's prediction of every subset of features given a dataset with m features.

A model retraining is required on all feature subsets $S \subseteq F$, where F are all the available features. A value of importance it is assigned to every variable that accounts for the impact on the model's prediction of incorporating that feature. A model $f_{S \cup \{i\}}$ is trained with that feature present and another model $f_S$ is trained with the feature withheld in order to compute this effect. Then, both models predictions are compared on the current input $f_{S \cup \{i\}}(x_{S \cup \{i\}}) - f_S(x_S)$, where xs are the values of the input variables in the set S. Since the effect of withholding a feature depends on other features in the model, the preceding differences are computed for all possible subsets $S \subseteq F \setminus \{i\}$. The feature attributions are the computed Shapley values.

They are a weighted average of all possible subsets of S in F:

$$\phi_i = \sum_{S \subseteq F \setminus \{i\}} \frac{|S|!(M-|S|-1)!}{M!} [f_{S \cup \{i\}}(x_{S \cup \{i\}}) - f_S(x_S)] \quad (5)$$

SHAP value is the only possible locally accurate and consistent feature contribution values [62], [63], they can provide high quality explanation both local and global.

Calculating the importance of the features based on SHAP contributions, the mean of each feature is retrieved for each SHAP matrix. Then, the resulting vectors are summed.

### b. SHAP Dependence Plots

For every feature and data instance, a point is plotted with the feature value on the x-axis and the corresponding Shapley value on the y-axis, this is the SHAP feature dependence plot.

Mathematically, the plot contains the following points:

$$\left\{ \left( x_j^{(i)}, \phi_j^i \right) \right\}_{i=1}^n \quad (6)$$

SHAP dependence plots are an alternative to partial dependence plots and accumulated local effects. While other methods show average effects, SHAP dependence also shows the variance on the y-axis.

### c. Rules Extraction

Tree ensembles such as random forests and boosted trees are accurate but difficult to understand. In this work, the framework of the interpretable tree (inTrees) is used to extract, measure, prune, select, and summarize rules from a tree ensemble and calculate frequent variable interactions[64].

Tree ensemble methods consist of multiple decision trees [53], [55]. A rule can be extracted by means of a decision tree's root node to a leaf node.

This rule summarization process explained at algorithm 3, is relevant in order to understand and filter the rules for phenomenon interpretability.

Given a rule {C ⇒ T}, where C is the condition's rule, being a conjunction of variable-value pairs aggregated from the path from the root node to the current node, $C_{node}$ denote the variable-value pair used to split the current node, $leaf_{Node}$ denote the flag whether the current node is a lead node, $pred_{node}$ denote the prediction at a leaf node, and T for rule's output.

| ALGORITHM III. ruleExtract algorithm |
|---|
| **Precondition:** |
| • **Input:** $ruleSet \leftarrow null, node \leftarrow rootNode, C \leftarrow null$ |
| • **Ouput:** $ruleSet$ |
| 1    **function:** $ruleExtract(\boldsymbol{ruleSet}, \boldsymbol{node}, \boldsymbol{C})$ |
| 2        **if** $leaf_{Node}$ = *true* **then** |
| 3            $currentRule \leftarrow \{C \rightarrow pred_{node}\}$ |
| 4            $ruleSet \leftarrow \{ruleSet \rightarrow currentRule\}$ |
| 5            **return** *ruleset* |
| 6        **end if** |
| 7        **for** $child_i = every\ child\ of\ node$ **do**: |
| 8            $C \leftarrow C \wedge C_{node}$ |
| 9            $ruleSet \leftarrow ruleExtract(ruleSet, child, C)$ |
| 10      **end for** |
| 11      **return** *ruleSet* |
| 12   **end function** |

The method ruleExtract explained at pseudocode Algorithm 3 shows the method used to extract rules from a decision tree. As tree ensembles are multiple decision trees, the final rules are a combination of rules extracted from each decision tree in the tree ensemble.

In the following work, it is applied the inTrees framework to the data set. For the winning classifier, the ruleExtract method is applied. As a result, several rules are extracted, and a post-processing rules step is performed. This post-processing comprises de-duping rules and rules metrics computation for rules quality. The rule's metrics are length which is the number of conditions within a rule, support which is the percentual frequency of observations that fulfil the rule, the rule's error for classification tasks which is the number of correctly classified instances within a rule condition and the target lift (epigraph II.B.1) for every rule as the number proportion of positive targets in the rule condition compared with the variable range.

## II. RESULTS & DISCUSSION

In this work, a methodology is proposed for understanding the economic recession phenomenon and extracting rules as an early economic recession detection method with a balance of getting a model with a suitable accuracy for prediction, which is the main scope of interpretable models in machine learning. This methodology begins with benchmarking proposed models to get the feature importance for the winning model (see epigraph II.C.4.a). From this step, the main variables that lever the economic recession are detected by understanding the dependencies with the most correlated variables and the feature value interaction regarding the target variable to understand this phenomenon better (see epigraph II.C.4.b). To conclude, a rule extraction process is performed for proposing rules useful for early detection of economic recession (see epigraph II.C.4.c).

As the first step, two tree-based classification models are fitted to the data; as a result, Table IV shows the results for the proposed accuracy metrics for the fitted models. When assessing the predictive accuracy, the yield curve performs quite well. Additional information can improve its predictive performance [65]. Thus, the main purpose of this work is through term spreads as unique independent variables to build a model for interpretability with a balance on predictive accuracy.

TABLE IV. Classification metrics results.

| Model | Class | Precision | Recall | Specificity |
|---|---|---|---|---|
| RF | 0 | 0.88 | 0.96 | 0.25 |
|  | 1 | 0.52 | 0.25 | 0.96 |
| XGB | 0 | 0.96 | 1.00 | 0.80 |
|  | 1 | 1.00 | 0.80 | 1.00 |

Despite adding only variables about interest rate nature, suitable classification metrics are obtained employing term spread variables for predicting an economic recession. XGB model has better classification metrics results; for the positive target class, the precision shows us how no false positives are obtained; for this reason, specificity also has the maximum value. However, recall has a high value but not the maximum, showing that despite a balanced classification of negative and positive labels, false negatives are present. After fitting and selecting the winning model, the model interpretation for understanding the phenomenon as the most important part of this work comes with the feature importance as the first relevant output to interpret which variables are the main predictors for economic recessions. The variable importance is obtained by computing the mean of absolute SHAP value for all instances for every feature at the training and test set. As a result, Table V, which is in the appendix, is plotted in Fig.3 for better understanding. In Fig.3, the features are sorted by variable importance in descending order from top to bottom for the most relevant and less relevant, respectively. Besides, by only considering the presence of variables Fig.3.A and 3.B shows similar results at the most important variables; however, as the test set has the more recent data, it is expected to be more representative for future values and may be more accurate in order to extrapolate this information for a near future, due to this, the main analysis is focused in the test set analysis.

information about the economy, so the particular choices regarding the maturity amount mainly to fine-tuning process.

In previous studies, the best results are obtained when forecasting an economic recession by taking the difference between two interest rates whose maturities are far apart.[65] suggested that the 3-month–10-year term spread provides a suitable combination of accuracy and validity in the long term to predict economic recessions. However, most term spreads are highly correlated and provide similar information about the economy, so the particular choices regarding the maturity amount mainly to fine-tuning process.

Results suggest that the most important term spreads are 3-month–6-month, 2-year–5-year, 5-year–10-year, 3-year–7-year, 3-year–3-month and 2-year–6-month. Although this work has more recent data than previous studies, the literature suggests as a rule of thumb that the difference between 10-year and 3-month Treasury rates becomes negative in early recessions providing a reasonable accuracy and time prevalence [65]. Despite not having this term spread as the more relevant, most term spreads are highly correlated and provide similar information about the economy's behaviour, so the particular choices concerning maturity amount mainly to fine-tuning and not to reversal of results[65]. The cautionary is that a reference point that works for one spread may not work for others. For example, the 2-year to 10-year term spread may reverse in advance of the 3-month to 10-year term spread, which tends to be higher[1]. In this line, some of the most critical variables like 5-year - 10-year term spread align with the literature statements as could invert earlier than 10-year–3-month term spread.

SHAP contribution values are plotted for training and test sets in Fig.4.A and 4.B. This method estimates an individual sample because they are local explainers. Nonetheless, this can lead to different results as training and test set have different instances; in this case, there are slight differences between both results. Besides, this plot retrieves additional information about the feature value analysis and the position of the instances on the plot. The horizontal location shows whether the effect of that value is associated with a higher or lower prediction from right to left; respectively, the vertical location shows the variable importance. The colour gradient shows whether that variable is high (dark) or low (light) for that observation.

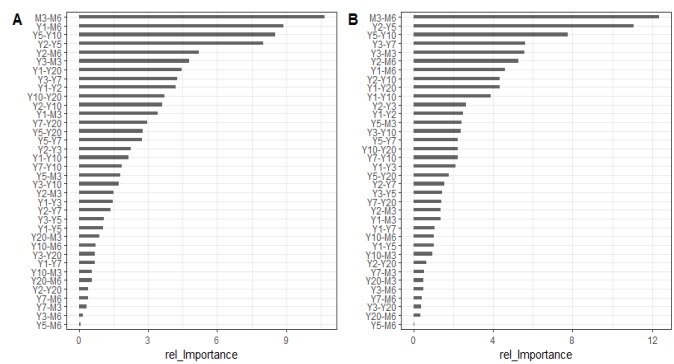

Fig. 3. Training(A) &Test(B) SHAP values for the variables.

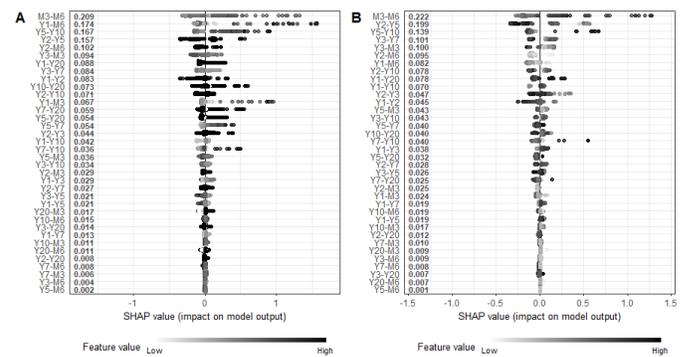

Fig. 4. Training(A) & Test(B) SHAP contribution values results.

In previous studies, the best results are obtained when forecasting an economic recession by taking the difference between two interest rates whose maturities are far apart.[65] suggested that the 3-month–10-year term spread provides a suitable combination of accuracy and validity in the long term to predict economic recessions. However, most term spreads are highly correlated and provide similar

As argued before, the analysis is focused on test set results, SHAP contribution values analysis could be complementary to decile target lift results at Table II as it is a preliminary analysis that has not the best splitting method for finding a range with the maximum split. SHAP contribution analysis shows that 3-year–6-month and 5-year–10-year term spreads have a higher lift for higher values, the 9-10 deciles. The term mentioned above spreads shows this relationship information at the SHAP contribution plot at Fig.4, the dark gradient colour for instances are at the right side of the plot and the light ones at the left, which indicates that high values are associated with positive predictions of economic recession. On the contrary, an opposite behaviour is shown on 2-year–5-year, 3-year–7-year, 3-year–3-month and 2-year–6-year spreads, which is somehow aligned with the decile target lift values of Table II, the lower values, the higher lift, in other words, higher probability of economic recession. As the SHAP contribution plot shows local interpretability and the decile target lift is not an optimized method for splitting ranges for maximizing lift, these complementary results also may present different nuances at both results due to are different perspective analyses.

Once the main features that impact economic recession prediction are detected, the dependent variables with more important variables on the target variable are studied. Dependence plots have been explained at epigraph II.C.4.b; more information can be found at [62], [63]. In essence, this plot shows feature values of the most important variables on the x-axis and SHAP values of the most correlated variable on the y-axis; additionally, a gradient colour to the points by the feature value of the designated variable is added.

For selecting the most correlated variable, the pairwise Pearson's correlation is performed at subsection II.B. By sorting the correlation coefficient, the most important variable is selected as the most correlated feature; as a result, Table VI at the appendix. Results suggest that the most correlated variables for the most important ones are in the same time term; for long term time spreads, most correlated are long term ones. The relevance of this information is to complement the previous findings with the dependencies of other variables to know the dependence and relationship between the most important variables and the most correlated to them; this helps complete the overview of the processes that affect the economic recessions.

The dependence plot for the most important variables is shown in Fig.5. At the x-axis, the horizontal location is the actual value from the most correlated variable, and at the y-axis, the vertical location shows what having that value did to the prediction. Additionally, the relationship between both information is shown with a loess regression line. For positive slopes, this trend says that the more variable value, the higher the model's prediction is for the most correlated variable; it is the opposite with negative slopes.

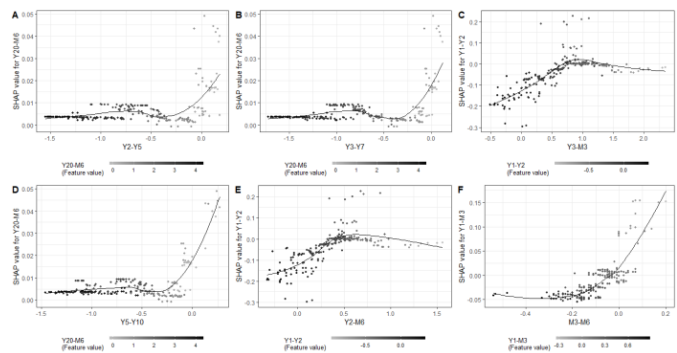

Fig. 5. SHAP dependence plot for most important variables and their most correlated features.

As a result, two kinds of relationships are found: one with a positive trend at Fig.5 plots A, B, D and F with a positive slope, having the highest correlation with 20-year–6month and 1-year–3-months term spread respectively. Besides, the positive trend with an asymptotic behaviour at Fig.5 plots C and E is found to correlate a 1-year–2-year term spread. In addition, the colour gradient shows the y-axis feature value from light to dark when variables value is low to high, respectively. Generally speaking, the more considerable value of the most correlated variables, the smaller the SHAP value of this variable is. At this point, decile target lift, feature contribution, feature importance and feature dependence are presented; this information let understanding as early indicators which initial range variable values have more probability of having economic recessions and which variable are the most relevant for the economic recession process respectively.

To finalize, at epigraph II.C.4.c is proposed a methodology for identifying rules for economic recession detection. As a result of rules extraction and initial postprocessing, 359 rules are extracted followed by rules metrics; due to saving space, and the table is not presented in the appendix; this can be asked in a document enclosed to this work.

The extracted rules from the winning model can be filtered in several ways; as an initial exploratory study, this work proposes a frequency maximization and Lift Maximization criterion for discovering interesting rules. Frequency maximization criterion is when rules are sorted by support in descending order, and the first rules are the most frequent. The frequency maximization criterion does not sort results by lift, error or length metric for the rules.

TABLE VII. Top 5 XGB Max support Rules.

| Rule | Error | Length | Support | Lift |
|---|---|---|---|---|
| M3-M6 ≤ 0.19 | 0.14 | 1 | 0.95 | 1.02 |
| Y5-Y10 ≤ 0.35 | 0.12 | 1 | 0.95 | 0.84 |
| Y2-Y5 ≤ 0.15 | 0.13 | 1 | 0.90 | 0.95 |
| Y3-M3 > 0.26 | 0.12 | 1 | 0.85 | 0.90 |
| Y3-Y7 ≤ -0.1 | 0.09 | 1 | 0.74 | 0.59 |

[a] Source is in an enclosed document, rules are obtained by sorting by support and selecting by the presence of top variables from SHAP results.
[b] M and Y are referred for monthly term and yearly respectively.

Table VII shows some rules for the Frequency maximization criterion, and results show a maximum Support for a rule of 0.95% of observations that satisfy the condition. By analyzing lift criterion, these rules show values nearly to 1, which is equivalent to saying that these rules could guarantee that there is no special probability of finding an economic recession compared with other data range; however, a rule with values near to 0 could show a high probability of not finding an economic recession. As previously explained, XGB is a tree-ensemble model through assembling simple trees, making a complex non-linear model. In this way, the rules extraction may provide rules with a low level of complexity. Due to this sorting method, the most important rules present low Length, low Lift and error rate, qualifying these as simplistic and inaccurate rules.

By sorting rules by lift in descending order, the first rules impact the economic recession detection more. Nevertheless, these rules could affect little observations, but as a recession is a rare event, support for recession identification should be a small percentage.

TABLE VIII. Top 5 XGB Max lift and support Rules.

| Rule | Error | Length | Support | Lift |
|---|---|---|---|---|
| Y2-M6≤-0.145 & Y20-M3>0.79 | 0 | 2 | 0.01 | 7.17 |
| Y2-Y3 ≤ -0.12 & Y5-Y10 ≤ 0.04 & M3-M6 > 0.01 | 0 | 3 | 0.03 | 6.32 |
| Y1-Y2 > -0.585 & Y2-M6 > 1.02 | 0 | 2 | 0.04 | 6.32 |
| Y5-Y10 > 0.12 & Y5-Y20 ≤ 0.43 & Y20-M6 > -0.66 | 0 | 3 | 0.04 | 6.32 |
| Y3-M3 > 0.45 & Y5-Y20 > 0.22 & | 0 | 3 | 0.04 | 6.32 |

Y5-M3 ≤ 1.32

<sup>a</sup> Source is in an enclosed document, rules are obtained by lift and selecting by the presence of top variables from SHAP results.
<sup>b</sup> M and Y are referred for monthly term and yearly respectively.

Table VIII shows some rules for lift maximization criterion; results show a maximum Lift for a rule of 7.17 times more probability of economic recession for the observations that satisfy the condition comparing the overall observations. Nonetheless, as an economic recession is a rare event, these rules usually have low support due to the nature of the economic recession, which is a rare event. More complex rules are found by this sorting criterion, with a low error rate and high probability of economic recession; therefore, the more interesting rules may be found. The interpretation of these rules is pretty straightforward, and a condition value is presented for every term spread involved in the rule; when this condition is satisfied, support, the percentage of observation that satisfies this rule is computed with the respective lift.

For the first rule, 2-year–6-month and 20-year–3-month are involved; this also indicates an interaction in the rule between these variables regarding the economic recession detection. Besides, the 20-year–the 3-month term spread is also an important term spread indicator as it may invert earlier than the 3-month–the 10-year term spread stated as relevant in previous studies [65].

Regarding the threshold values interpretation, the values are compared with the min, mean and max values for all the historical data for every term spread(see Table IX at appendix) in order to interpret the threshold value as a small, average or big value as those thresholds are closer to any of this feature descriptive statistics, in the case a value is close to two statistics the priority for the average is given. As a result, the first threshold number is labelled as a small value and the second as an average value. In this way, the qualitative interpretation of this rule will be formulated as follows: "When the 2-year–6-month term spread is lower or equal a small value and 20-year–3-month term spread is greater than the average value there is over seven times more probability of economic recession than the probability of economic recession for the complementary conditions". Besides, historically this rule fulfilled the economic recessions accounted for 2008.

For the second rule, 2-year–3-year, 5-year–10-year and 3-month–6-month are involved, mainly describing an interaction between these variables regarding the economic recession detection. "When the Y2–Y3 and Y5–Y10 term spread is lower or equal of the average value of this term spread and greater than the average value of M3–M6 term spread, there is over six times more probability of economic recession than the probability of economic recession for the complementary conditions". Besides, these conditions were fulfilled in the economic recessions accounted at 1990, 1991, 2001 and 2008.

The other rules from Table VIII can be described similarly to the previously explained rules, and these rules fulfil the conditions of the economic recession accounted at 1980, 1981, 1982, 1974 & 1970 years. This technique allows us to have a set of rules for detecting economic recession; with proper data updating & model retraining, these rules can be used in real life and act consequently with economic policies, among other uses.

To summarize the findings, Table X shows the main results except for dependencies analysis results.

TABLE X. Summary Table of empirical results

| Variables | Most Correlated | Decile lift | SHAP(+) | Rules Support | Rules Lift |
|---|---|---|---|---|---|
| M3-M6 | Y1-M3 | Low-High | High | ✓ | ✓ |
| Y2-Y5 | Y20-M6 | Mid-High | Low-Mid | ✓ | ✗ |
| Y5-Y10 | Y20-M6 | Mid-High | High | ✓ | ✓ |
| Y3-Y7 | Y20-M6 | Mid-High | Low-Mid | ✓ | ✗ |
| Y3-M3 | Y1-Y2 | Low-Mid | Low-Mid | ✓ | ✓ |
| Y2-M6 | Y1-Y2 | Low-High | Low | ✗ | ✓ |

As a result, main variables on predicting economic recession are detected, and the variable dependence concerning the most correlated is studied; the SHAP value for positive economic recession is taken into account with the preliminary information of Decile Target Lift. Besides, some of the top rules contain the most important variables and fulfil the ideas mentioned in this work.

III. CONCLUSION

Regarding the term structure, long-term rates could explain changes in future short-term rates. Understanding the term structure and yield curve, our goal is to create an interpretable forecasting model that can accurately inform us about future recessions, which could be a valuable tool for practitioners, researchers, governments and central banks. For three main groups, the public sector and the private sector are households, banks and investors, and the Federal Reserve. From an investors point of view, this information could be useful to make the right decisions for investing considering different strategies regarding this information, as the expanding economic activity is correlated with the stock market expansion[66]. By using the term spread to know in advance a possible economic recession, Federal Reserve could modify the interest rates to try to reduce the effect of this phenomenon.

Relevant term spreads are found, 3-month - 6-month, 2-year–5-year, 5-year–10-year, 3-year–7-year, 3-year–3-month and 2-year–6-month. Furthermore, for these variables, the lift metric is computed in order to detect initial intervals with a higher probability of accounting for a recession which is complementary to the SHAP contribution values analysis, applied into the rules description methods implementing the necessary policy mix they can dampen the effects of the recession, minimize its duration, or steer the economy away from it altogether. As the model provides some false negative alarms, we expect that implementing fiscal and monetary policy may put some inflationary pressure on the economy.

Finally, the methodology proposes a novelty application in this topic by extracting rules for economic recession understanding and detection. With this technique, several descriptive conditions allow the user to understand this phenomenon and have indicators with the goal of detecting to minimize the magnitude of the effect of the recession.

It is important to note that the yield curve's predictive power is statistical evidence and that, despite its accuracy, it is impossible to assure future results.

Thus, we encourage validating and updating these rules with reasonable frequency as the market evolves.

The literature suggests that the USA's best predictor of economic recessions is the 3-month-10-year term spread. Nevertheless, we found that the 3-month-6-month spread is the most relevant for detecting recessions, including the main recession detection rules. Therefore, monitoring this spread can be a useful tool for recession identification and a valid indicator for market expectations. In this context, it is found that the best rule associates this short-term 3-month-6-month predictor with the long-term term spreads, such as

5-year-10-year and 2-year-3-year, illustrating the rule as "When the Y2-Y3 and Y5-Y10 term spread is lower or equal of the average value of this term spread and greater than the average value of M3-M6 term spread there is over six times more probability of economic recession than the probability of economic recession for the complementary conditions".

As a future work suggestion, several paths can be followed. On one accuracy side, the improvement of the model predictive accuracy is relevant to have tools with high quality and impact on predicting this phenomenon. On the interpretability side, as different exogenous variables can be added, more study on the variable interactions can be performed to understand the yield curve inversion with other variables relevant for generating policies to prevent and control. On the rules generation side, as rules are potentially changing over time as variable importance may variate, a predictive maintenance system could be proposed to keep rules updated and valid over time.

APPENDIX

TABLE V. SHAP values for train and test set

| Feature | Training | Test |
|---|---|---|
| M3-M6 | 0.2095 | 0.2217 |
| Y2-Y5 | 0.1571 | 0.1986 |
| Y5-Y10 | 0.1674 | 0.1394 |
| Y3-Y7 | 0.0837 | 0.1012 |
| Y3-M3 | 0.0939 | 0.1002 |
| Y2-M6 | 0.1022 | 0.0949 |
| Y1-M6 | 0.1745 | 0.0824 |
| Y1-Y20 | 0.0877 | 0.0778 |
| Y2-Y10 | 0.071 | 0.0778 |
| Y1-Y10 | 0.0422 | 0.0699 |
| Y2-Y3 | 0.0442 | 0.0474 |
| Y1-Y2 | 0.0827 | 0.0445 |
| Y5-M3 | 0.0355 | 0.0432 |
| Y3-Y10 | 0.0337 | 0.0431 |
| Y10-Y20 | 0.0731 | 0.0403 |
| Y5-Y7 | 0.0536 | 0.0403 |
| Y7-Y10 | 0.0363 | 0.0403 |
| Y1-Y3 | 0.0292 | 0.0381 |
| Y5-Y20 | 0.0545 | 0.0321 |
| Y2-Y7 | 0.0272 | 0.0278 |
| Y3-Y5 | 0.0213 | 0.0262 |
| Y7-Y20 | 0.0585 | 0.025 |
| Y2-M3 | 0.0294 | 0.0248 |
| Y1-M3 | 0.0675 | 0.0244 |
| Y1-Y7 | 0.0134 | 0.0194 |
| Y10-M6 | 0.0145 | 0.0188 |
| Y1-Y5 | 0.0209 | 0.0186 |
| Y10-M3 | 0.011 | 0.017 |
| Y2-Y20 | 0.0081 | 0.0116 |
| Y7-M3 | 0.0065 | 0.0096 |
| Y20-M3 | 0.0173 | 0.0093 |
| Y3-M6 | 0.0036 | 0.0087 |
| Y7-M6 | 0.0076 | 0.0078 |
| Y3-Y20 | 0.0137 | 0.0067 |
| Y20-M6 | 0.0108 | 0.0066 |
| Y5-M6 | 0.0015 | 0.0007 |

[a] Term spread are sorted by SHAP values in percent scale of test set.

TABLE VI. Pearson correlation coefficient for the most correlated variable.

| Variable | Correlated | Correlation |
|---|---|---|
| Y1-Y10 | Y20-M6 | -0,98 |
| Y20-M6 | Y1-Y10 | -0,98 |
| Y1-Y7 | Y20-M6 | -0,97 |
| Y1-Y5 | Y10-M6 | -0,97 |
| Y10-M6 | Y1-Y5 | -0,97 |
| Y1-Y20 | Y20-M6 | -0,97 |
| Y7-M6 | Y1-Y5 | -0,97 |
| Y2-Y5 | Y20-M6 | -0,96 |
| Y20-M3 | Y1-Y7 | -0,96 |
| Y2-Y7 | Y20-M6 | -0,96 |
| Y10-M3 | Y1-Y5 | -0,96 |
| Y1-Y3 | Y7-M6 | -0,96 |
| Y5-M6 | Y1-Y3 | -0,96 |
| Y7-M3 | Y1-Y3 | -0,95 |
| Y1-Y2 | Y5-M6 | -0,94 |
| Y2-Y10 | Y20-M6 | -0,94 |
| Y2-Y3 | Y20-M6 | -0,94 |
| Y5-M3 | Y1-Y3 | -0,94 |
| Y3-Y5 | Y20-M6 | -0,93 |
| Y3-Y7 | Y20-M6 | -0,93 |
| Y3-M6 | Y1-Y2 | -0,92 |
| Y2-Y20 | Y20-M6 | -0,91 |
| Y3-Y10 | Y20-M6 | -0,90 |
| Y3-M3 | Y1-Y2 | -0,90 |
| Y5-Y7 | Y20-M6 | -0,87 |
| Y3-Y20 | Y20-M6 | -0,87 |
| Y5-Y10 | Y20-M6 | -0,83 |
| Y2-M6 | Y1-Y2 | -0,81 |
| Y5-Y20 | Y20-M6 | -0,80 |
| Y2-M3 | Y1-Y2 | -0,79 |
| Y1-M3 | M3-M6 | -0,75 |
| M3-M6 | Y1-M3 | -0,75 |
| Y7-Y20 | Y20-M6 | -0,73 |
| Y7-Y10 | Y20-M6 | -0,73 |
| Y10-Y20 | Y20-M6 | -0,65 |
| Y1-M6 | M3-M6 | -0,44 |

TABLE IX. Term spread descriptive statistics

| Feature | mean | median | min | max | sd |
|---|---|---|---|---|---|
| Y1-Y2 | -0.29 | -0.31 | -1.06 | 0.95 | 0.34 |
| Y1-Y3 | -0.46 | -0.51 | -1.63 | 1.77 | 0.55 |
| Y1-Y5 | -0.75 | -0.77 | -2.50 | 2.35 | 0.81 |
| Y1-Y7 | -0.98 | -1.02 | -2.87 | 2.82 | 0.99 |
| Y1-Y10 | -1.14 | -1.18 | -3.40 | 3.07 | 1.15 |
| Y1-Y20 | -1.42 | -1.33 | -4.15 | 3.33 | 1.38 |
| Y1-M3 | 0.52 | 0.43 | -0.94 | 2.93 | 0.44 |
| Y1-M6 | 0.39 | 0.31 | -0.39 | 1.60 | 0.32 |
| Y2-Y3 | -0.17 | -0.17 | -0.59 | 0.83 | 0.22 |
| Y2-Y5 | -0.49 | -0.46 | -1.55 | 1.41 | 0.53 |
| Y2-Y7 | -0.74 | -0.71 | -2.28 | 1.88 | 0.72 |
| Y2-Y10 | -0.93 | -0.85 | -2.83 | 2.13 | 0.91 |
| Y2-Y20 | -1.30 | -1.14 | -3.67 | 2.39 | 1.19 |
| Y2-M3 | 0.79 | 0.72 | -1.76 | 3.86 | 0.66 |
| Y2-M6 | 0.69 | 0.61 | -0.82 | 2.44 | 0.54 |
| Y3-Y5 | -0.30 | -0.27 | -0.99 | 0.58 | 0.31 |
| Y3-Y7 | -0.53 | -0.50 | -1.72 | 1.05 | 0.52 |
| Y3-Y10 | -0.69 | -0.60 | -2.36 | 1.30 | 0.71 |
| Y3-Y20 | -1.01 | -0.82 | -3.27 | 1.56 | 1.00 |
| Y3-M3 | 0.97 | 0.98 | -2.01 | 4.11 | 0.80 |
| Y3-M6 | 0.85 | 0.86 | -1.20 | 2.74 | 0.69 |
| Y5-Y7 | -0.23 | -0.21 | -0.76 | 0.47 | 0.22 |
| Y5-Y10 | -0.39 | -0.31 | -1.46 | 0.72 | 0.42 |
| Y5-Y20 | -0.73 | -0.60 | -2.47 | 1.25 | 0.72 |
| Y5-M3 | 1.27 | 1.33 | -2.25 | 4.33 | 0.99 |
| Y5-M6 | 1.15 | 1.20 | -1.56 | 3.12 | 0.89 |
| Y7-Y10 | -0.16 | -0.11 | -0.74 | 0.38 | 0.22 |
| Y7-Y20 | -0.50 | -0.42 | -1.80 | 0.84 | 0.52 |
| Y7-M3 | 1.50 | 1.56 | -2.49 | 4.46 | 1.12 |
| Y7-M6 | 1.37 | 1.43 | -2.03 | 3.31 | 1.03 |
| Y10-Y20 | -0.34 | -0.34 | -1.06 | 0.87 | 0.34 |
| Y10-M3 | 1.66 | 1.74 | -2.65 | 4.42 | 1.24 |
| Y10-M6 | 1.53 | 1.59 | -2.28 | 3.64 | 1.16 |
| Y20-M3 | 1.93 | 2.01 | -3.00 | 4.44 | 1.41 |
| Y20-M6 | 1.80 | 1.79 | -2.54 | 4.36 | 1.35 |
| M3-M6 | -0.12 | -0.10 | -1.45 | 1.01 | 0.19 |

## REFERENCES

[1] Estrella, A. (2005a). The yield curve as a leading indicator: frequently asked questions. New York Fed.
[2] Holmes, M. J., Otero, J., & Panagiotidis, T. (2015). The expectations hypothesis and decoupling of short-and long-term US interest rates: A pairwise approach. The North American Journal of Economics and Finance, 34, 301-313.
[3] Liu, W., & Moench, E. (2016). What predicts US recessions?. International Journal of Forecasting, 32(4), 1138-1150.
[4] Shiller, R. J. (1990). The term structure of interest rates. Handbook of monetary economics, 1, 627-722.
[5] Estrella, A., & Hardouvelis, G. A. (1991). The term structure as a predictor of real economic activity. Journal of Finance, 46, 555-576.
[6] Weber, E., & Wolters, J. (2012). The US term structure and central bank policy. Applied Economic Letters, 19, 41-45.
[7] Weber, E., & Wolters, J. (2013). Risk and policy shocks on the us term structure. Scottish Journal of Political Economy, 60, 101-119.
[8] Campbell, J. Y. (1995). Some lessons from the yield curve. Journal of economic perspectives, 9(3), 129-152.
[9] Bernanke, B. S., & Blinder, A. S. (1992). The federal funds rate and the channels of monetary transmission. The American Economic Review, 901-921.
[10] Vides, J. C., Iglesias, J., & Golpe, A. A. (2018). The Term Structure Under Non-linearity Assumptions: New Methods in Time Series. In New Methods in Fixed Income Modeling (pp. 117-136). Springer, Cham.
[11] Vetzal, K. R. (1994). A survey of stochastic continuous time models of the term structure of interest rates. Insurance: Mathematics and Economics, 14(2), 139-161.
[12] Estrella, A., & Trubin, M. (2006). The yield curve as a leading indicator: Some practical issues. Current issues in Economics and Finance, 12(5).
[13] Estrella, A., & Mishkin, F. S. (1996). The yield curve as a predictor of US recessions. Current issues in economics and finance, 2(7).
[14] Poole, W., Rasche, R. H., & Thornton, D. L. (2002). Market anticipations of monetary policy actions. Review-Federal Reserve Bank of Saint Louis, 84(4), 65-94.
[15] Estrella, A., & Hardouvelis, G. (1990). Possible roles of the yield curve in monetary policy. Federal Reserve Bank of New York, Intermediate Targets and Indicators for Monetary Policy-A Critical Survey, July, 339-362.
[16] Dueker, M. J. (1997). Strengthening the case for the yield curve as a predictor of US recessions. Federal Reserve Bank of St. Louis Review, 41-51.
[17] Laurent, R. D. (1988). An interest rate-based indicator of monetary policy. Economic Perspectives, (Jan), 3-14.
[18] Harvey, C. R. (1989). Forecasts of economic growth from the bond and stock markets. Financial Analysts Journal, 45(5), 38-45.
[19] Stock, J. H., & Watson, M. W. (1989). New indexes of coincident and leading economic indicators. NBER macroeconomics annual, 4, 351-394.
[20] Chen, N. F. (1991). Financial investment opportunities and the macroeconomy. The Journal of Finance, 46(2), 529-554.
[21] Harvey, C. R. (1993). Term structure forecasts economic growth. Financial Analysts Journal, 49(3), 6-8.
[22] Dotsey, M. (1998). The predictive content of the interest rate term spread for future economic growth. FRB Richmond Economic Quarterly, 84(3), 31-51.
[23] Hamilton, J. D., & Kim, D. H. (2002). A re-examination of the predictability of the yield spread for real economic activity. Journal of Money, Credit, and Banking, 34(2), 340-360.
[24] Estrella, A. (2005b). Why does the yield curve predict output and inflation?. The Economic Journal, 115(505), 722-744.
[25] Ang, A., Piazzesi, M., & Wei, M. (2006). What does the yield curve tell us about GDP growth?. Journal of econometrics, 131(1-2), 359-403.
[26] Wheelock, D. C., & Wohar, M. E. (2009). Can the term spread predict output growth and recessions? A survey of the literature. Federal Reserve Bank of St. Louis Review, 91(5 Part 1), 419-440.
[27] Estrella, A., Rodrigues, A. P., & Schich, S. (2003). How stable is the predictive power of the yield curve? Evidence from Germany and the United States. Review of Economics and Statistics, 85(3), 629-644.
[28] Rudebusch, G. D., & Williams, J. C. (2009). Forecasting recessions: the puzzle of the enduring power of the yield curve. Journal of Business & Economic Statistics, 27(4), 492-503.
[29] Nyberg, H. (2010). Dynamic probit models and financial variables in recession forecasting. Journal of Forecasting, 29(1-2), 215-230.
[30] Lahiri, K., Monokroussos, G., & Zhao, Y. (2013). The yield spread puzzle and the information content of SPF forecasts. Economics Letters, 118(1), 219-221.
[31] Chinn, M., & Kucko, K. (2015). The predictive power of the yield curve across countries and time. International Finance, 18(2), 129-156.
[32] Evgenidis, A., Tsagkanos, A., & Siriopoulos, C. (2017). Towards an

asymmetric long run equilibrium between stock market uncertainty and the yield spread. A threshold vector error correction approach. Research in International Business and Finance, 39, 267-279.

[33] Gebka, B., & Wohar, M. E. (2018). The predictive power of the yield spread for future economic expansions: Evidence from a new approach. Economic Modelling, 75, 181-195.

[34] Evgenidis, A., Papadamou, S., & Siriopoulos, C. (2020). The yield spread's ability to forecast economic activity: What have we learned after 30 years of studies?. Journal of Business Research, 106, 221-232.

[35] Ng, S. (2017). Opportunities and challenges: Lessons from analyzing terabytes of scanner data (No. w23673). National Bureau of Economic Research.

[36] Berge, T. J. (2015). Predicting recessions with leading indicators: Model averaging and selection over the business cycle. Journal of Forecasting, 34(6), 455-471.

[37] Gogas, P., Papadimitriou, T., Matthaiou, M., & Chrysanthidou, E. (2015). Yield curve and recession forecasting in a machine learning framework. Computational Economics, 45(4), 635-645.

[38] Gogas, P., Papadimitriou, T., & Chrysanthidou, E. (2015). Yield curve point triplets in recession forecasting. International Finance, 18(2), 207-226.

[39] Döpke, J., Fritsche, U., & Pierdzioch, C. (2017). Predicting recessions with boosted regression trees. International Journal of Forecasting, 33(4), 745-759.

[40] Vrontos, S. D., Galakis, J., & Vrontos, I. D. (2020). Modeling and predicting US recessions using machine learning techniques. International Journal of Forecasting.

[41] Bluwstein, K., Buckmann, M., Joseph, A., Kang, M., Kapadia, S., & Şimşek, Ö. (2020). Staff Working Paper No. 848 Credit growth, the yield curve and financial crisis prediction: evidence from a machine learning approach.

[42] Wei, P., Lu, Z., & Song, J. (2015). Variable importance analysis: a comprehensive review. Reliability Engineering & System Safety, 142, 399-432.

[43] Yun, Y. H., Deng, B. C., Cao, D. S., Wang, W. T., & Liang, Y. Z. (2016). Variable importance analysis based on rank aggregation with applications in metabolomics for biomarker discovery. Analytica chimica acta, 911, 27-34.

[44] Yang, S., Tian, W., Heo, Y., Meng, Q., & Wei, L. (2015). Variable importance analysis for urban building energy assessment in the presence of correlated factors. Procedia Engineering, 121, 277-284.

[45] Vladislavleva, E., Friedrich, T., Neumann, F., & Wagner, M. (2013). Predicting the energy output of wind farms based on weather data: Important variables and their correlation. Renewable energy, 50, 236-243.

[46] Dohoo, I., Ducrot, C., Fourichon, C., Donald, A. and Hurnik, D. (1997), "An overview of techniques for dealing with large numbers of independent variables in epidemiologic studies", Preventive Veterinary Medicine, Vol. 29 No. 3, pp. 221-239.

[47] Tufféry, S. (2011). Data mining and statistics for decision making. John Wiley & Sons.

[48] Ferreira, A. J., & Figueiredo, M. A. (2012). Boosting algorithms: A review of methods, theory, and applications. In Ensemble machine learning (pp. 35-85). Springer, Boston, MA.

[49] Sun, Q., & Pfahringer, B. (2011, December). Bagging ensemble selection. In Australasian Joint Conference on Artificial Intelligence (pp. 251-260). Springer, Berlin, Heidelberg.

[50] Ho, T. K. (1995, August). Random decision forests. In Proceedings of 3rd international conference on document analysis and recognition (Vol. 1, pp. 278-282). IEEE.

[51] Kleinberg, E. M. (1990). Stochastic discrimination. Annals of Mathematics and Artificial intelligence, 1(1-4), 207-239.

[52] Kleinberg, E. M. (2000). On the algorithmic implementation of stochastic discrimination. IEEE Transactions on Pattern Analysis and Machine Intelligence, 22(5), 473-490.

[53] Breiman, L. (2001). Random forests. Machine learning, 45(1), 5-32.

[54] Wright, M. N., & Ziegler, A. (2015). ranger: A fast implementation of random forests for high dimensional data in C++ and R. arXiv preprint arXiv:1508.04409.

[55] Friedman, J. H. (2001). Greedy function approximation: a gradient boosting machine. Annals of statistics, 1189-1232.

[56] Tianqi Chen and Carlos Guestrin, "XGBoost: A Scalable Tree Boosting System", 22nd SIGKDD Conference on Knowledge Discovery and Data Mining, 2016, https://arxiv.org/abs/1603.02754

[57] Otte, C. (2013). Safe and interpretable machine learning: A methodological review. In Computational intelligence in intelligent data analysis (pp. 111-122). Springer, Berlin, Heidelberg.

[58] Wei, P., Lu, Z., & Song, J. (2015). Variable importance analysis: a comprehensive review. Reliability Engineering & System Safety, 142, 399-432.

[59] Strobl, C., Boulesteix, A. L., Zeileis, A., & Hothorn, T. (2007). Bias in random forest variable importance measures: Illustrations, sources and a solution. BMC bioinformatics, 8(1), 25.

[60] Strobl, C., Boulesteix, A. L., Kneib, T., Augustin, T., & Zeileis, A. (2008). Conditional variable importance for random forests. BMC bioinformatics, 9(1), 307

[61] Lundberg, S. M., Erion, G., Chen, H., DeGrave, A., Prutkin, J. M., Nair, B., ... & Lee, S. I. (2020). From local explanations to global understanding with explainable AI for trees. Nature machine intelligence, 2(1), 2522-5839.

[62] Shapley regression values: Lipovetsky, Stan, and Michael Conklin. "Analysis of regression in game theory approach." Applied Stochastic Models in Business and Industry 17.4 (2001): 319-330.

[63] Lundberg, S. M., & Lee, S. I. (2017). A unified approach to interpreting model predictions. In Advances in neural information processing systems (pp. 4765-4774).

[64] Deng, H. (2019). Interpreting tree ensembles with intrees. International Journal of Data Science and Analytics, 7(4), 277-287.

[65] Estrella, A., & Mishkin, F. S. (1998). Predicting US recessions: Financial variables as leading indicators. Review of Economics and Statistics, 80(1), 45-61.

[66] Cameron, D. R. (1978). The expansion of the public economy: A comparative analysis. The American Political Science Review, 1243-1261.

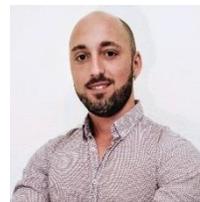

### Pedro Cadahía Delgado

Principal Data Scientist PepsiCo. PhD Computer Science student in Economics. He has worked as Data Scientist in different fields and firms such as: DB Schenker, Minsait by Indra, Schibsted (Infojobs), Cofidis and PepsiCo.

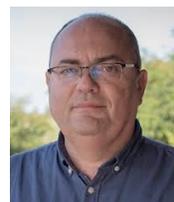

### Emilio Congregado

Emilio Congregado is Full Professor in the Economics Department at the University of Huelva. His main research interests are in the area of Entrepreneurship and labor economics. His work has been published in several scientific journals including Small Business Economics, Journal of Business Venturing, Empirical Economics, Journal of Policy Modeling, Annals of Regional Science or Economic Modeling.

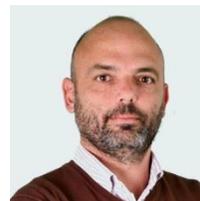

### Antonio Golpe Moya

Antonio A. Golpe is Associate Professor in the Economics Department at University of Huelva. His main research interests are in the area of Applied Economics. His work has been published in several scientific journals including Energy Economics, Annals of Regional Science, Renewable and Sustainable Energy Review, International Small Business Journal, Empirical Economics, Journal of Policy Modeling, Economic Modeling or International Journal of Tourism Research.

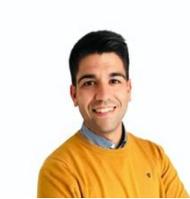

### Jose Carlos Vides

Jose Carlos Vides is Assistant Professor in the Department of Applied and Structural Economics and History at Complutense University of Madrid. His main research interests are in the area of Applied Economics. His work has been published in several scientific journals including Energy Economics, International Review of Economics and Finance, Journal of Policy Modeling, Empirica, Tourism Economics or FinanzArchiv: Public Finance Analysis.